\documentclass[iicol,sn-apa]{sn-jnl}% referee option is meant for double line spacing

\usepackage[utf8]{inputenc} % allow utf-8 input
\usepackage[T1]{fontenc}    % use 8-bit T1 fonts
\usepackage{hyperref}       % hyperlinks
\usepackage{url}            % simple URL typesetting
\usepackage{booktabs}       % professional-quality tables
\usepackage{amsfonts}       % blackboard math symbols
\usepackage{nicefrac}       % compact symbols for 1/2, etc.
\usepackage{microtype}      % microtypography
\usepackage[table]{xcolor}
\usepackage{times}
\usepackage{epsfig}
\usepackage{graphicx}
\usepackage{amsmath}
\usepackage{amssymb}
\usepackage{subfiles}
\usepackage{makecell} 
\usepackage{wrapfig} 
\usepackage{bm}
\usepackage{subfig}
\usepackage{tabularx}
\usepackage{enumitem}
\usepackage{bbm}  
\usepackage{xspace}
\usepackage{multirow}
\usepackage{caption}
\usepackage{threeparttable}
\usepackage[capitalize]{cleveref}
\crefname{section}{Sec.}{Secs.}
\Crefname{section}{Section}{Sections}
\Crefname{table}{Table}{Tables}
\crefname{table}{Tab.}{Tabs.}

\makeatletter
\DeclareRobustCommand\onedot{\futurelet\@let@token\@onedot}
\def\@onedot{\ifx\@let@token.\else.\null\fi\xspace}
\def\eg{\emph{e.g}\onedot} 

\def\ie{\emph{i.e}\onedot}

\def\vs{\emph{vs}\onedot}

\def\etal{\emph{et al}\onedot}
\makeatother

\definecolor{mygray}{gray}{.95}
\definecolor{mylightergray}{gray}{.99}
\definecolor{mygreen}{RGB}{10, 179, 33}
\makeatletter
\newcommand{\thickhline}{%
    \noalign {\ifnum 0=`}\fi \hrule height 1pt
    \futurelet \reserved@a \@xhline
}
\newcolumntype{"}{@{\vrule width 1pt}}

\newcommand{\revision}[1]{\textcolor{black}{{#1}}}

\newcommand{\R}{\mathbb{R}}
\newcommand{\encoder}{\mathcal{E}(\cdot)}
\newcommand{\decoder}{\mathcal{D}(\cdot)}
\newcommand{\backbone}{\Phi(\cdot)}
\newcommand{\videofeat}{\mathbf{F}}
\newcommand{\encact}{\tilde{\mathbf{E}}^{\text{act}}}
\newcommand{\encpos}{\tilde{\mathbf{E}}^{\text{pos}}}
\newcommand{\encnum}{{L}^{\text{enc}}}
\newcommand{\decnum}{{L}^{\text{dec}}}
\newcommand{\encacthead}{\tilde{\mathbf{A}}}
\newcommand{\encposehead}{\tilde{\mathbf{P}}}
\newcommand{\selectact}{\mathbf{E}^{\text{act}}}
\newcommand{\selectpos}{\mathbf{E}^{\text{pos}}}
\newcommand{\queryact}{\mathbf{Q}^{\text{act}}}
\newcommand{\querypos}{\mathbf{Q}^{\text{pos}}}
\newcommand{\decact}{\mathbf{D}^{\text{act}}}
\newcommand{\decpos}{\mathbf{D}^{\text{pos}}}
\newcommand{\predpos}{\mathbf{P}}
\newcommand{\predact}{\mathbf{A}}

\newcommand{\gtpos}{\hat{\mathbf{P}}}
\newcommand{\gtact}{\hat{\mathbf{A}}}

\newcommand{\nullcls}{\emptyset}

\newcommand{\gtval}{\hat{N}}
\newcommand{\midpoint}{\mathbf{m}}
\newcommand{\duration}{\mathbf{d}}
\newcommand{\permq}{\mathfrak{S}_Q}
\newcommand{\match}{\hat{\sigma}}

\newcommand{\gtall}{\hat{\mathbf{Y}}}
\newcommand{\predall}{\mathbf{Y}}

\newcommand{\lossmatch}{\mathcal{L}_{\text{match}}}
\newcommand{\lossiou}{\mathcal{L}_{\text{gIoU}}}
\newcommand{\lambdaiou}{\lambda_{\text{gIoU}}}
\newcommand{\lambdahug}{\lambda_{\text{Hungarian}}}
\newcommand{\lambdall}{\lambda_{\text{L1}}}

\newcommand{\losspos}{\mathcal{L}_{\text{pos}}}
\newcommand{\losshug}{\mathcal{L}_{\text{Hungarian}}}
\newcommand{\lossctrs}{\mathcal{L}_{\text{ctrs}}}
\newcommand{\lambdactrs}{\lambda_{\text{ctrs}}}

\newcommand{\github}{\href{https://shirleymaxx.github.io/DeTRC/}{project page}\xspace}

\begin{document}

\title[Efficient Action Counting with Dynamic Queries]{Efficient Action Counting with Dynamic Queries}

\author[1]{\fnm{Xiaoxuan} 
\sur{Ma}}\email{maxiaoxuan@pku.edu.cn}

\author[1]{\fnm{Zishi} \sur{Li}}\email{mrblack\_lizs@outlook.com}

\author[1]{\fnm{Qiuyan} \sur{Shang}}\email{shangqiuyan@stu.pku.edu.cn}

\author[1]{\fnm{Wentao} \sur{Zhu}}\email{wtzhu@pku.edu.cn}

\author[1]{\fnm{Hai} \sur{Ci}}\email{cihai@pku.edu.cn}

\author[2]{\fnm{Yu} \sur{Qiao}}\email{qiaoyu@sjtu.edu.cn}

\author*[1]{\fnm{Yizhou} \sur{Wang}}\email{yizhou.wang@pku.edu.cn}

\affil[1]{\orgdiv{School of Computer Science}, \orgname{Peking University}, \orgaddress{\city{Beijing}, \postcode{100871}, \country{China}}}

\affil[2]{\orgdiv{School of Electronic Information and Electrical Engineering}, \orgname{Shanghai Jiao Tong University}, \orgaddress{\city{Shanghai}, \postcode{200240},  \country{China}}}


\abstract{
Most existing methods rely on the similarity correlation matrix to characterize the repetitiveness of actions, but their scalability is hindered due to the quadratic computational complexity. In this work, we introduce a novel approach that employs an action query representation to localize class-agnostic repeated action cycles with linear computational complexity. Based on this representation, we develop two key components to tackle the essential challenges of temporal repetition counting. Firstly, to tackle open-set action counting, \revision{we define two action classes: ``repetitive actions'' and ``others''. Instead of manually defining the repetitive action class, we propose a dynamic action query strategy. Here, each action query directly represents an extracted video feature, allowing the repetitive actions of interest to be dynamically defined based on the video content itself. Secondly, to distinguish these repetitive action queries from others, we propose inter-query contrastive learning. This performs contrastive clustering over the queries, pulling similar action patterns together while pushing apart those related to background or unrelated movements. As a result, queries classified as ``repetitive actions'' are considered as repetitive cycles, which are then used for counting. Thanks to the query-based representation and contrastive learning strategy, our method significantly outperforms previous works on accuracy while being more lightweight and time-efficient.} On the challenging RepCountA benchmark, we outperform the state-of-the-art method TransRAC by 26.5\% in OBO accuracy, with a 22.7\% mean error decrease and 94.1\% computational burden reduction. \revision{Code and models are publicly available at \github.}
}

\keywords{Temporal repetition counting, Video understanding}



\maketitle

\section{Introduction}
\label{sec:intro}

Temporal periodicity is a ubiquitous phenomenon in the natural world. Temporal Repetition Counting (TRC) aims to accurately measure the number of repetitive action cycles within a given video and holds significant potential for applications such as fitness monitoring~\citep{fieraru2021aifit} and motion generation \citep{zhu2023human}. 

\begin{figure*}[t]
    \centering
    \includegraphics[width=\linewidth]{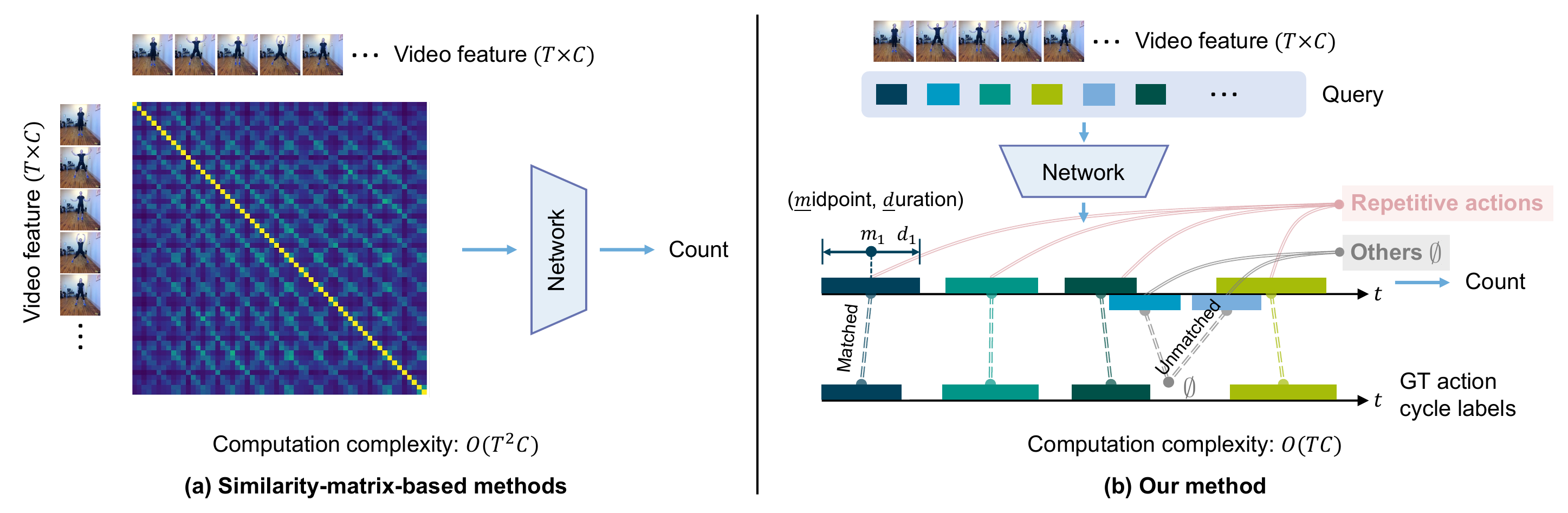}
    \vspace{0.05cm}
    \caption{\textbf{Conceptual workflow comparison of (a) similarity-matrix-based methods and (b) our proposed action query-based method.} (a) Most existing methods use similarity matrices calculated between each frame to detect repetitive actions, resulting in a time complexity of $O(T^2C)$, where $T$ denotes video length and $C$ denotes feature dimension. (b) On the other hand, our method employs action queries to represent each action cycle and estimates their classes and temporal locations. Each query is classified into either ``repetitive actions'' or ``others ($\nullcls$)'' class labels. The temporal location is expressed by the midpoint and duration on the timeline. During training, we perform bipartite matching to uniquely associate a prediction to a GT action cycle, taking into account both class labels and temporal locations. Predictions with no match should yield a $\nullcls$ class prediction, indicating that these are not repetitive actions. This novel formulation reduces the complexity from quadratic to linear as $O(TC)$.}
    \label{fig:structure}
\end{figure*}

Pioneer methods ~\citep{laptev2005periodic,azy2008segmentation,cutler2000robust,tsai1994cyclic,pogalin2008visual,thangali2005periodic,chetverikov2006motion} represent time-series video data as one-dimensional signals and employ spectral analysis techniques such as the Fourier transform. While suitable for short videos with fixed periodic cycle lengths, these methods struggle to handle real-world scenarios with varying cycle lengths and sudden interruptions. Recent studies shift to deep learning-based methods~\citep{levy2015live,dwibedi2020counting,hu2022transrac,zhang2020context,li2024repetitive} and show promising performances. 
Notably, most of these methods, such as RepNet~\citep{dwibedi2020counting} and TransRAC~\citep{hu2022transrac}, utilize a temporal similarity correlation matrix to depict repetitiveness, as illustrated in \cref{fig:structure} (a).
Nevertheless, the computational complexity of this representation grows quadratically with the number of input frames $T$, highlighting a significant gap in scalability that hinders their application to real-world scenarios of varying action periods and dynamics.

Recent progress in action detection \citep{shi2022react,liu2022end} introduces an efficient representation of action periods by associating each action instance with an action query, similar to DETR~\citep{carion2020end}. Inspired by this, we propose to formulate the TRC problem as a set prediction task where the goal is to detect every action cycle by representing it as an action query, as illustrated in \cref{fig:structure} (b). Based on this query representation, we use a Transformer encoder-decoder network \citep{zhu2020deformable} to detect repetitive action instances and their temporal positions, defined by their midpoints and durations. This novel formulation reduces the complexity from quadratic to linear \footnote{In implementation, we employ deformable attention modules as proposed in DeformableDETR \citep{zhu2020deformable}.} and enables counting long videos with varying action periods.
However, directly applying the action detection approaches \citep{liu2022end, zhang2022actionformer, shi2022react} to the TRC problem proves inadequate (\cref{tab:result-repcountA}) in addressing two distinctive challenges unique to TRC. These challenges underscore the complexity of TRC, highlighting why TRC is not merely another action detection task but requires a nuanced approach that considers the unique nature of repetitive actions. We highlight the two inherent differences between TRC and the classical action detection task:

\begin{enumerate}

    \item TRC requires recognizing \emph{open-set} action instances depending on the input video, rather than detecting predefined action classes in the detection task.
    
    \item TRC requires recognizing action instances with \emph{identical} content, while detection does not.
\end{enumerate}

As a result, approaching TRC as a simple action detection task results in inferior performance as shown in \cref{sec:sota}. In contrast, we propose two novel strategies to address these challenges and redefine the framework for TRC. 
In response to the first open-set challenge, \revision{we define two action classes: ``repetitive actions'' and ``others ($\nullcls$)''. Instead of manually defining the repetitive action class, we propose a \emph{Dynamic Action Query} (DAQ) strategy, which adaptively updates the action query using content features extracted from the video (\cref{fig:pipeline}) This mechanism allows the decoder to attend to the repetitive actions based on the input video contents in a dynamic, contextually aware and class-agnostic manner.}  
To tackle the second challenge, \revision{we further propose \emph{Inter-query Contrastive Learning} (ICL), which ensures that primary repetitive action cycles of interest are grouped together in the learned representation space while being separated from other distractors, such as background noise.}
The integration of two core components (DAQ and ICL) ensures that the action instances are identified adaptively based on the video content and their contextual similarity\revision{, making our approach class-agnostic.} In other words, our queries are designed to localize the contextually similar action instances, which aligns exactly with the definition of repetition counting. 
Extensive experiments validate the effectiveness of the two proposed designs.

We summarize our contributions as follows:
\begin{enumerate}
    \item We provide a novel perspective to tackle the TRC problem using a simple yet effective representation for action cycles, \ie action query. Our approach reduces the computational complexity from quadratic to linear and is \revision{more lightweight and time-efficient}.
    \item We propose \emph{Dynamic Action Query} to guide the model to focus on the repetitive actions contextually defined by the video content thereby improving generalization ability across different actions.
    \item We introduce \emph{Inter-query Contrastive Learning} to facilitate learning primary repetitive action representations and to distinguish them from distractions.
    \item Our method notably surpasses state-of-the-art (SOTA) methods in terms of both accuracy and efficiency on two challenging benchmarks. Notably, our method strikes an effective balance in handling various action periods and video lengths, offering a significant leap forward in the practical application of TRC technologies.
\end{enumerate}

\section{Related Work}
\label{sec:related-work}

\subsection{Temporal Repetition Counting}

Traditional methods~\citep{laptev2005periodic, azy2008segmentation, cutler2000robust, tsai1994cyclic, pogalin2008visual, thangali2005periodic, chetverikov2006motion} frequently employ spectral or frequency domain techniques for the analysis of repetitive sequences, thereby preserving the underlying repetitive motion structures. While these conventional approaches are capable of effectively handling simple motion sequences or those characterized by fixed periodicity, they prove inadequate when confronted with non-stationary motion sequences encountered in real-world scenarios. In contrast, deep-learning-based approaches~\citep{levy2015live, dwibedi2020counting, hu2022transrac, zhang2020context,li2024repetitive} have demonstrated remarkable performance improvements. Notably, RepNet~\citep{dwibedi2020counting} and TransRAC~\citep{hu2022transrac} leverage temporal similarity matrices of actions to construct models for counting temporal repetitions. However, these similarity-matrix-based methods are not scalable for long videos due to their quadratic computational complexity. Another research line involves predicting the start and end points of each cycle~\citep{zhang2020context} from coarse to fine. Nevertheless, its practicality is hindered by the requirement for over 30 forward passes to count iteratively from a single video. In this paper, we introduce an effective action cycle representation by leveraging a Transformer encoder-decoder, which reduces the computational complexity from quadratic to linear and demonstrates superior performance in handling both fast and slow actions.

\subsection{Temporal Action Detection}

The field of temporal action detection~\citep{redmon2016you, zhao2017temporal, chao2018rethinking, zhang2022actionformer, lin2019bmn} is typically classified into two categories: anchor-based methods, and anchor-free methods. Anchor-based methods~\citep{zeng2019graph,li2021three,qing2021temporal} generate multiple anchors, subsequently classifying these anchors to determine the action boundaries. Anchor-free methods ~\citep{buch2019end, shou2017cdc, yuan2017temporal,lin2021learning} predict action instances by directly regressing the boundary and the center point of an action instance. 
With the rapid development of Transformer technology, DETR~\citep{carion2020end} is introduced for object detection task~\citep{zhu2020deformable,meng2021conditional,liu2022dab,zhang2022dino} and gains increasing popularity with promising performance. This paradigm promotes the study in many fields such as the action detection tasks~\citep{liu2022end, tan2021relaxed, vaswani2017attention, wang2021oadtr}. These methods establish a direct connection between action queries and the predicted action instances, enabling them to accurately predict the temporal boundaries of actions. Inspired by these promising results, we explore the possibility of utilizing a novel action query to represent the action cycle in TRC task. In contrast to existing action detection methods, our approach allows the model to capture the inherent repetitive content of an action cycle without relying on predefined class labels and effectively addresses confounding factors such as non-repetitive video backgrounds. This makes our approach well-suited for tackling the challenges of the TRC problem.

\section{Method}
\label{sec:methods}

\subsection{Preliminary}
\label{subsec:detr}

DETR \citep{carion2020end} is a pioneering object detection framework that builds upon the Transformer encoder-decoder architecture \citep{vaswani2017attention}. The overall DETR architecture \citep{carion2020end} consists of three main components: a backbone to extract image features, an encoder-decoder Transformer, and the detection heads, \ie feed-forward network (FFN) that makes the final detection prediction. The main features of DETR are the conjunction of a bipartite matching loss and transformers with (non-autoregressive) parallel decoding. The bipartite matching loss is a set-based Hungarian loss during training, which uniquely assigns a prediction to a GT object, and is invariant to a permutation of predicted objects. This design enables DETR to perform parallel processing and predict all objects simultaneously. We briefly review the workflow as follows.

Given the input image feature maps extracted by a CNN backbone, \eg ResNet \citep{he2016deep}, DETR exploits a standard Transformer encoder-decoder network to transform the feature maps to be features of a set of object queries. An FFN and a linear projection are added on top of the object query features as the detection heads. The FFN acts as the regression branch that predicts the bounding box coordinates, \ie box center coordinates, box height and width. The linear projection acts as the classification branch to produce the classification results, \ie $\mathtt{object}$ \vs $\mathtt{no\, object}\, (\nullcls)$. The $\nullcls$ class is used to represent that no object is detected, playing a similar role to the ``background'' class in the standard object detection approaches.

DETR infers a fixed-size set of $N$ predictions in a single pass, where $N$ is set to be significantly larger than the typical number of objects in an image. During training, a Hungarian loss produces an optimal bipartite matching between predicted and GT objects and then optimizes the object position-specific losses. The matching procedure takes into account both the class prediction and the similarity of predicted and GT boxes and finds one-to-one matching for direct set prediction without duplicates.

\begin{figure*}[t]
    \centering
    \includegraphics[width=\linewidth]{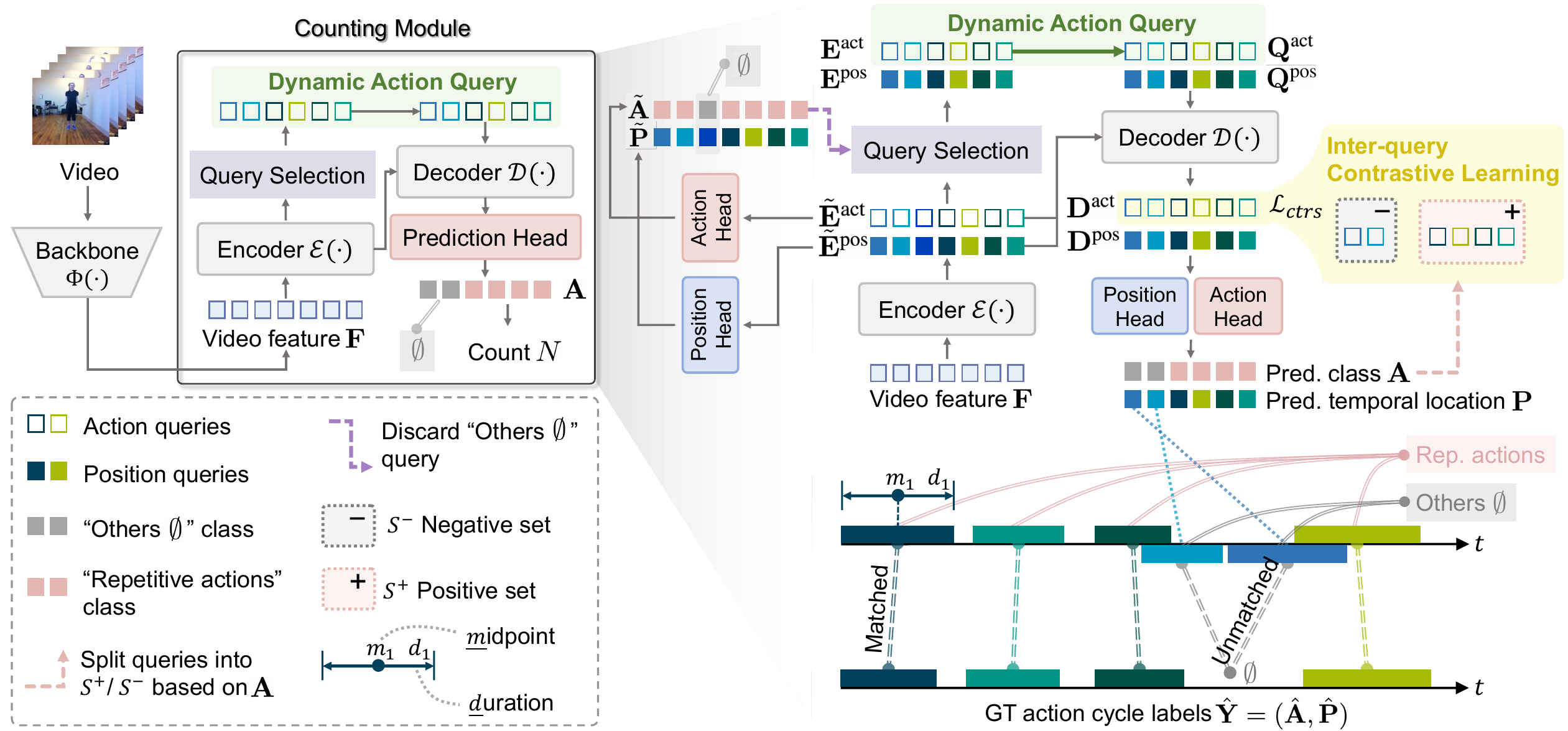}
    \vspace{0.05cm}
    \captionsetup{labelfont={color=black}}\caption{\textbf{Overview and detailed architecture design of our method.} \textbf{(Left)} We utilize a DETR-inspired framework to predict the number of repetitive action cycles given a video input. \revision{The framework consists of a backbone network $\backbone$ for feature extraction and the counting module (black box). The counting module, including an encoder-decoder Transformer for processing query features and prediction heads for action classification, outputs queries classified as ``repetitive actions'', which are then counted to yield the total count $N$.} \textbf{(Right)} The detailed workflow of \revision{the counting module.} Given the video features $\videofeat$ as input, the encoder $\encoder$ produces \revision{a set of query features. A query selection module then screens these, retaining only the most relevant ones for forwarding to the decoder $\decoder$. 
    The decoder $\decoder$ adopts the Dynamic Action Query (DAQ) strategy, using the selected action queries for initialization to dynamically define the ``repetitive actions'' class. The output embeddings from the decoder are then passed to their corresponding prediction head, which estimates the class label $\predact$ and temporal location $\predpos$. During training, we employ bipartite matching to uniquely pair each prediction with a GT action cycle $\gtpos$. Predictions that fail to match are assigned the $\nullcls$ class. To further distinguish the ``repetitive actions'' from $\nullcls$ actions, we propose Inter-query Contrastive Learning (ICL). Finally, the queries classified as repetitive actions contribute to the total count $N$.}} 
    \label{fig:pipeline}
    \vspace{-0.3cm}
\end{figure*}

\subsection{Overview}
\label{subsec:overview}

Given an RGB video sequence with $T$ frames, the TRC task aims to predict an integer $N$ indicating the number of detected \revision{primary} repetitive action cycles\revision{, whose class is not predefined}. Drawing inspiration from DETR \citep{carion2020end}, we streamline the problem as an \revision{open-}set detection task in the temporal domain \revision{and propose to use the \textit{action query} to represent each potential repetitive cycle.} The overall framework consists of \revision{two main components: a backbone network $\backbone$ and the counting module. The backbone extracts video features $\videofeat$. The counting module, depicted as the black box in \cref{fig:pipeline}, is composed of a Transformer encoder $\encoder$, a decoder $\decoder$, and prediction heads. The extracted $\videofeat$ are transformed into a set of action queries by the Transformer encoder and decoder. Two prediction heads are then added on top of these query features, aiming to classify each query and estimate its temporal location. Once an action query is classified as a repetitive cycle, the action count $N$ increases by 1.}

\revision{Considering the two challenges of the TRC tasks discussed in \cref{sec:intro}, we propose \textbf{Dynamic Action Query (DAQ)} strategy to address the open-set problem. Specifically, we define two action classes: ``repetitive actions'' and the ``others ($\nullcls$)'' class. Instead of manually defining the specific class label for the ``repetitive actions'', DAQ directly leverages dynamically updated video features from the encoder to initialize the action query for the decoder. This enables the adaptive definition of ``repetitive actions'' based on the video content, in a contextually aware and class-agnostic manner. To tackle the second challenge of distinguishing \textit{identical} repetitive actions, we further propose \textbf{Inter-query Contrastive Learning (ICL)}. This approach clusters queries representing identical repetitive actions into a positive action set ($S^+$) and groups the other queries into a negative set ($S^-$) in the feature space. Integrating DAQ and ICL allows our method to identify class-agnostic similar action instances that are adaptively based on the video content, and exclude other distracting actions at the same time.}

Similar to DETR \citep{carion2020end}, during training, we employ bipartite matching which uniquely assigns a prediction to a GT action cycle. The matching procedure takes into account both the class prediction and the similarity of predicted and GT temporal locations. We then optimize the action cycle-specific losses. In inference, our method produces a fixed-size set of $Q$ predictions in a single pass, where $Q$ is set to be significantly larger than the typical number of action cycles in a $T$-frame video sequence. By counting the queries classified as ``repetitive actions'', we get the final total count value $N$.

In the following, we will introduce the network architecture design in \cref{subsec:arch}, the DAQ and ICL modules in \cref{subsec:DAQ} and \cref{subsec:ICL}, and the model training in \cref{subsec:loss}.

\subsection{Model Architecture}
\label{subsec:arch}
\revision{As illustrated in \cref{fig:pipeline}, our model can be divided into two main components: a backbone $\backbone$ and the counting module (black box). The backbone extracts features $\videofeat$ from the raw video input. The counting module then takes these video features $\videofeat$ as input and ultimately outputs the count N. Specifically, the counting module includes a Transformer encoder-decoder, a query selection module, prediction heads, and a bipartite matching module. Next, we introduce each of them in detail. \\}

\noindent\textbf{Backbone.}
The backbone network $\backbone$ takes a sequence of $T$ video frames as input and extracts feature vectors $\mathbf{F} \in \R^{T \times C}$ for each frame, where $C$ denotes the feature dimension.  \\

\noindent\textbf{Encoder.} The encoder $\mathcal{E}(\cdot)$ is a classical Transformer \citep{vaswani2017attention} architecture which has $\encnum$ standard encoder layers. The encoder transforms the video features $\videofeat$ into two distinct query types: \textit{action queries} $\encact \in \R^{T \times C}$ and \textit{position queries} $\encpos \in \R^{T \times C}$, as shown in \cref{fig:pipeline} (right). \revision{Action queries capture features pertinent to action classification, while position queries concentrate on the temporal dimensions of an action. These features prepare the decoder for action query initialization, from which it will identify repetitive action cycles.} Please refer to the supplementary material for the detailed encoder architecture. \\

\noindent\textbf{Query selection.}
\label{subsec:query}
We add a query selection module before passing the encoder output query features to the decoder, as depicted on the right side of \cref{fig:pipeline}. \revision{This module aims to perform an initial filtering over the encoded video features $\videofeat$. Since the encoder $\encoder$ produces a large number of tokens $\encact \in \R^{T \times C}$ and $\encpos \in \R^{T \times C}$, many of them may correspond to background or distracting content. Therefore the query selection module selects a subset of $Q$ informative features from the original $T$ tokens as $\selectact \in \R^{Q \times C}$ and $\selectpos \in \R^{Q \times C}$. In practice, we} route the query features outputted by the encoder to two respective prediction heads (described later in this section), which independently decode them into predictions for action class and temporal location. The action head processes the action queries $\encact$ to estimate the action class $\encacthead$ for each query, while the position head decodes the position queries $\encpos$ to their temporal locations $\encposehead$. Based on the prediction results, we discard query features classified as ``others ($\nullcls$)'', retaining only those identified as ``repetitive actions''. For the remaining queries, we rank them based on their classification confidence and preserve only the top $Q$ high-confidence queries. The others are discarded. After this selection process, both the action query features $\selectact$ and position query features $\selectpos$ have a dimension of $Q \times C$. Additional details will be provided when we discuss the prediction head later in this section. \\

\noindent\textbf{Decoder.} The decoder $\mathcal{D}(\cdot)$ is also a classical Transformer \citep{vaswani2017attention} architecture, consisting of $\decnum$ standard decoder layers. It processes a set of action queries $\queryact \in \R^{Q \times C}$ and position queries $\querypos \in \R^{Q \times C}$, while simultaneously attending to the direct output (\ie $\encact$ and $\encpos$) from the encoder, \revision{as depicted in \cref{fig:pipeline} (right)}. Following the practice in DETR \citep{carion2020end}, these inputs are decoded in parallel across each decoder layer and transformed into the corresponding query features $\decact \in \R^{Q \times C}$ and $\decpos \in \R^{Q \times C}$. The input position queries $\querypos$ are initialized as learnable parameters. To address the open-set challenge in the TRC task, we introduce the DAQ strategy, which initializes the decoder input action queries using the selected encoder embeddings $\selectact$, \ie $\queryact = \selectact$. The details of the DAQ strategy will be elaborated in \cref{subsec:DAQ}. Using self- and encoder-decoder attention over these embeddings, our method globally reasons selected video frame features together while being able to use the whole frame-wise video feature as context. The two sets of $Q$ queries are then independently transformed into corresponding prediction results, \ie an action head decodes $\decact$ into the action class prediction $\predact$, and a position head decodes $\decpos$ into the temporal location prediction $\predpos$. We detail the two heads in the following subsection. \\

\noindent\textbf{Prediction heads.} The prediction heads are designed to make the final predictions using the features processed by the encoder-decoder\revision{, as shown in \cref{fig:pipeline} (right)}. They comprise two networks: an action head and a position head, both of which are Multi-Layer Perceptrons (MLPs).

The action head processes the output action queries from the decoder $\decact$ to estimate the action class $\predact \in \R^{Q \times 1}$ for each query using a softmax function. Each element of $\predact$ represents the probability that the corresponding query is a ``repetitive action'', with values spanning from 0 to 1. In practice, a threshold value, $\alpha$, is used to categorize the queries: values greater than $\alpha$ classify a query as a ``repetitive action'' while values less than or equal to $\alpha$ categorize it as ``others ($\nullcls$)''. Recall that in the query selection module, we rank the queries based on their estimated probabilities $\encacthead \in \R^{T \times 1}$, which serve as confidence scores. We then select the top $Q$ queries with the highest confidence scores according to the ranking.

The position head estimates the temporal location $\predpos = (\midpoint, \duration) \in \R^{Q \times 2}$ for each query based on the output position queries from the decoder $\decpos$. We use the midpoint $\midpoint \in \R^Q$ and the duration $\duration \in \R^Q$ to denote the temporal location of an action cycle.   \\

\noindent\textbf{Bipartite matching.}
\revision{Note that our method produces a fixed-size set of $Q$ predictions in a random order, where $Q$ is set to be much larger than the typical number of action cycles in a $T$-frame video sequence. To enforce supervision, we employ a matching strategy to pair the predicted action cycles (classes and temporal locations) with the GT action cycles, as illustrated in the bottom right of \cref{fig:pipeline}.}

Let us denote GT repetitive action cycle labels by $\gtall = (\gtact, \gtpos)$. $\gtact = \mathbf{1} \in \R^{\gtval \times 1}$ denotes the $\gtval$ repetitive action class labels, where we assign the value 1 to the ``repetitive actions''. $\gtpos \in \R^{\gtval \times 2}$ denotes the GT temporal positions of the $\gtval$ repetitive action, represented by $\gtval$ sets of midpoints and duration. Similarly, we denote the predicted repetitive actions by $\predall = (\predact, \predpos)$, a set of size $Q$ predictions. We set $Q$ much larger than $\gtval$ empirically. We consider $\gtall$ also a set of size $Q$ padded with ``others ($\nullcls$)'', \ie we assign the value 0 to denote $\nullcls$ class. To find a bipartite matching between these two sets $\gtall$ and $\predall$, we search for a permutation of $Q$ elements $\sigma \in \permq$ with the lowest cost:
\setlength{\abovedisplayskip}{1pt} 
\setlength{\belowdisplayskip}{4pt}
\begin{equation}
\label{eq:matching}
\hat{\sigma} = \arg\min_{\sigma \in \permq} \sum_{i=1}^Q \lossmatch(\gtall_i, \predall_{\sigma(i)}),
\end{equation}
where $\lossmatch$ is a pair-wise \textit{matching cost} between GT action cycle $\gtall_i$ and a prediction cycle with index $\sigma(i)$. 

The matching cost takes into account both the action classification result and the similarity of the predicted temporal locations and the GT temporal location. Each element $i$ of the GT action cycle set can be seen as $\gtall_i = (\gtact_i, \gtpos_i)$ where $\gtact_i$ is the target class label (1 for ``repetitive actions'', and 0 for $\nullcls$) and $\gtpos_i \in \R^2$ is a vector that denotes the midpoint time position and lasting duration of a GT action cycle. For the prediction with index $\sigma(i)$, we define probability of class $\gtact_i$ as $p_{\sigma(i)}(\gtact_i)$. Then we can define the matching cost as
\begin{align}
\label{eq:matchingcost}
\begin{split}
\lossmatch(\gtall_i, \predall_{\sigma(i)}) 
&= - \mathbbm{1}_{\{\gtact_i \neq \nullcls\}} p_{\sigma(i)}(\gtact_i) \\
&+ \mathbbm{1}_{\{\gtact_i \neq \nullcls\}} \losspos(\gtpos_i, \predpos_{\sigma(i)}),
\end{split}
\end{align}
where $\mathbbm{1}$ is an indicator function.

To measure the similarity of the predicted temporal locations and the GT temporal location, we define $\losspos$ using the linear combination of the $L_1$ distance and the generalized Intersection over Union (IoU) loss \citep{rezatofighi2019generalized}. Overall, the position loss is defined as 
\begin{align}
\label{eq:losspos}
\begin{split}
\losspos(\gtpos_i, \predpos_{\sigma(i)}) 
&= \lambdall \| \gtpos_i - \predpos_{\sigma(i)} \|_1 \\
&+ \lambdaiou \lossiou(\gtpos_i, \predpos_{\sigma(i)}),
\end{split}
\end{align}
where $\lambdaiou, \lambdall \in \R$ are hyperparamters.

By employing the Hungarian matching algorithm \citep{kuhn1955hungarian} to optimize \cref{eq:matching}, we can achieve the final optimal matching $\match$ which uniquely assigns a prediction to a GT action cycle, finding one-to-one matching without duplicates. Notice that the matching cost between a repetitive action instance and $\nullcls$ doesn’t depend on the prediction, which means that in that case the cost is a constant. Following this step, we can apply the corresponding losses related to classification and temporal location prediction. 

We define a Hungarian loss for all pairs matched in the previous step to supervise both classification and temporal location predictions, \ie a linear combination of a negative log-likelihood loss for class prediction and a temporal position loss defined in \cref{eq:losspos}:
\begin{align}
\begin{split}
    \losshug(\gtall, \predall) 
    &= \sum\limits_{i=1}^{Q} \left[ -\text{log}\,{p_{\match(i)}}{(\gtact_i)} \right. \\
    &+ \left. \mathbbm{1}_{\{\gtact_i \neq \nullcls\}} \losspos(\gtpos_i, \predpos_{\match(i)}) \right],
\end{split}
\end{align}
where $\match$ is the optimal assignment computed in the previous bipartite matching step, \ie \cref{eq:matching}.

\subsection{Dynamic Action Query}
\label{subsec:DAQ}
As discussed in \cref{sec:intro}, the TRC problem requires recognizing \textit{open-set} action instances depending on the video content, where the action category is not predefined. Therefore, \revision{we formulate the problem as a binary classification task with two categories: ``repetitive actions'' and ``others ($\nullcls$)''. \textit{The key idea is that the class of “repetitive actions” is not fixed or manually defined. Rather, it is dynamically inferred based on the video content.} To achieve this,} we propose the \textit{Dynamic Action Query} strategy. \revision{As shown in \cref{fig:pipeline} (top right), DAQ adaptively updates the action query $\queryact$ by directly assigning it the selected encoder action query features $\selectact$, \ie $\queryact \leftarrow \selectact$.} \revision{These decoder action queries $\queryact$ embed the video content, enabling the decoder to dynamically define ``repetitive actions'' based on the input video content. The DAQ strategy, simple yet effective, eliminates the need for manually defining action categories and thereby enhances the model's generalization capability.}

\revision{Besides, we also explore several different methods for initializing the queries in the decoder $\decoder$ in the supplementary, confirming that the DAQ strategy is the most effective.}

\subsection{Inter-query Contrastive Learning}
\label{subsec:ICL}
Since the input video may contain other distractors such as the background motion\footnote{\revision{We assume that each video contains only a primary repetitive action type, which is also the case for the public dataset.}}, another unique challenge to the TRC task is to recognize action instances with \textit{identical} content. This requires that the actions of interest we classify exhibit similarity in their motion patterns, while other action queries should have dissimilar representations. To tackle this challenge, we propose \textit{Inter-query Contrastive Learning} to distinguish the action queries\revision{, as shown in \cref{fig:pipeline} (right). ICL encourages the model to pull together queries corresponding to the same repetitive actions in the feature space, while pushing apart those related to background or unrelated movements. Therefore, we partition the action queries decoded by the decoder $\decact$ into two categories and employ contrastive learning on them based on the classification predictions $\predact$.} Specifically, features classified as ``repetitive actions'' form the positive set $S^+$, while the other features form the negative set $S^-$. Then we apply contrastive learning using InfoNCE loss~\citep{he2020momentum} $\lossctrs$ over the representation space:
\begin{align}
\begin{split}
\lossctrs &= - \sum\limits_{i \in S^+}{\rm log} \left( \frac {\mathcal{L}_{+}} {\mathcal{L}_{+} + \mathcal{L}_{-}} \right), \\
\mathcal{L}_{+} &= \sum\limits_{s \in S^+,s \neq i}{\rm exp} (\decact_{i} \cdot \decact_{s}) / \tau, \\
\mathcal{L}_{-} &= \sum\limits_{s \in S^-}{\rm exp} (\decact_{i} \cdot \decact_{s}) / \tau,
\end{split}
\end{align}
where $\tau$ is the temperature parameter, and $\cdot$ denotes inner product.

\subsection{Training}
\label{subsec:loss}
% We train the whole network in a supervised way. 
We train our model in an end-to-end manner using the overall loss function:
\begin{align}
    \mathcal{L} = \lambdahug\losshug + \lambdactrs\lossctrs,
\end{align}
where $\lambdahug, \lambdactrs \in \R$ are the coefficients. Following DETR \citep{carion2020end}, we also found it helpful to use auxiliary losses in the decoder during training. Specifically, we add $\losshug$ loss on the predictions from the prediction heads right after the encoder \ie $\encacthead$ and $\encposehead$, and add prediction heads and $\losshug$ loss after each decoder layer. All prediction heads share their parameters. This ensures the model focuses on the correct features at each stage consistently, thereby accelerating convergence.

\section{Experiments}
\label{sec:experiments}

\subsection{Datasets and Metrics}

\noindent\textbf{RepCountA dataset} \citep{hu2022transrac} is currently the largest and most challenging benchmark for the video TRC task~\footnote{The RepCountB \citep{hu2022transrac} test subset is proprietary and not publicly available.}. It is primarily compiled from fitness videos on YouTube, including a wide range of fitness activities conducted in diverse settings, including homes, gyms, and outdoor environments. This dataset stands out due to its extensive video lengths, significant variations in the average motion cycle, and a higher number of repetitive cycles compared to prior datasets~\citep{dwibedi2020counting, runia2018real, levy2015live,zhang2020context}.
We use the start and end positions of each action instance to compute the GT label $\gtpos$ provided by the annotations. We train our model on the RepCountA train set and select the best model on the validation set. We report the evaluation results on the test set.

\begin{table*}[t]
    \centering
    \begin{minipage}[t]{1.0\linewidth} %
        \centering
        \caption{\textbf{Comparison to the state-of-the-arts on RepCountA \citep{hu2022transrac} dataset.} We compare with SOTA action recognition/segmentation methods (top block), TRC methods (second block), and action detection methods (third block with $\dagger$). We further report MAE and OBO metrics for \underline{s}hort-, \underline{m}edium-, and \underline{l}ong-period test actions. }
        \label{tab:result-repcountA}
        \setlength{\tabcolsep}{3pt}
        \resizebox{\linewidth}{!}{%
        \begin{tabular}{l | c | c c | c c | c c | c c}
         \thickhline
          & Backbone & ${\rm MAE}\downarrow$ & ${\rm OBO}\uparrow$ & ${\rm MAE_{s}}\downarrow$ & ${\rm OBO_{s}}\uparrow$ & ${\rm MAE_{m}}\downarrow$ & ${\rm OBO_{m}}\uparrow$ & ${\rm MAE_{l}}\downarrow$ & ${\rm OBO_{l}}\uparrow$ \\
         \thickhline
         X3D \citep{feichtenhofer2020x3d} & X3D & 0.9105 & 0.1059 & - & - & - & - & - & - \\
         TANet \citep{liu2021tam} & TANet & 0.6624 & 0.0993 & - & - & - & - & - & - \\
         VideoSwinT \citep{liu2022video} & ViT & 0.5756 & 0.1324 & - & - & - & - & - & - \\
         GTRM \citep{huang2020improving} & I3D & 0.5267 & 0.1589 & - & - & - & - & - & - \\
         
         \thickhline
          RepNet \citep{dwibedi2020counting} & ResNet & 0.5865 & 0.2450 & 0.7793 & 0.0930 & 0.5893 & 0.1591 & 0.4549 & 0.4062 \\
          Zhang \etal \citep{zhang2020context} & 3D-ResNext & 0.8786 & 0.1554 & - & - & - & - & - & - \\
          TransRAC \citep{hu2022transrac} & ViT & 0.4891 & 0.2781 & 0.5789 & 0.0233 & 0.4696 & 0.2955 & 0.4420 & 0.4375 \\
          Li \etal \citep{li2024repetitive} & ViT & 0.3841 & 0.3860 & - & - & - & - & - & - \\
          \thickhline
          TadTR \citep{liu2022end} $\dagger$ & I3D & 1.1314 & 0.0662 & 0.8364 & 0.0233 & 1.1591 & 0.0000 & 1.3106 & 0.1406 \\
          ActionFormer \citep{zhang2022actionformer} $\dagger$ & I3D & 0.4990 & 0.2781 & 0.4164 & 0.1628 & 0.3768 & 0.3409 & 0.6385 & 0.3125 \\
          ReAct~\citep{shi2022react} $\dagger$ & TSN & 0.4592 & 0.3509 & 0.2805 & 0.1576 & 0.3037 & 0.4318 & 0.6862 & 0.3906 \\ 
          \thickhline
          \rowcolor{mygray}
          Ours & TSN & \underline{0.2809} & \underline{0.4570} & \underline{0.2411} & {0.1628} & \textbf{0.1792} & \underline{0.5455} & \underline{0.3776} & \underline{0.5938} \\
          \rowcolor{mygray}
          Ours & I3D & 0.3305 & 0.4437 & 0.2421 & \underline{0.2093} & 0.2012 & 0.5227 & 0.4788 & 0.5469 \\
          \rowcolor{mygray}
          Ours & ViT & \textbf{0.2622} & \textbf{0.5430} & \textbf{0.2257} & \textbf{0.2558} & \underline{0.2002} & \textbf{0.5909} & \textbf{0.3294} & \textbf{0.7031} \\
         \thickhline
        \end{tabular}}
    \end{minipage}
\end{table*}

\vspace{1em}
\noindent\textbf{UCFRep dataset} \citep{zhang2020context} is a subset of the UCF101 dataset~\citep{soomro2012ucf101}, including fitness videos and daily life videos. Following previous work \citep{hu2022transrac,li2024repetitive}, we do not use the train set but directly test our model on the test set to evaluate the model generalization ability.

\vspace{1em}
\noindent\textbf{Metrics.}
Following previous works \citep{hu2022transrac, dwibedi2020counting, zhang2020context, li2024repetitive}, we compute two commonly used metrics, OBO and MAE, to evaluate the model performance. \textbf{OBO} (Off-By-One count error) measures the percentage that the predicted count is within the GT count $\pm 1$ range. \textbf{MAE} (Mean Absolute Error) measures the normalized absolute difference between the predicted and GT counts. Formally,
\begin{align}
    {\rm OBO} &= \frac{1} {M} \sum_{m=1}^{M}\left| N_m - \hat{N}_m \leq 1 \right|, \\
    {\rm MAE} &= \frac{1} {M} \sum_{m=1}^{M} \frac{\left| N_m - \hat{N}_m\right|} {N_m},
\end{align}
where $M$ is the total number of test videos, ${N}_m$ and $\hat{N}_m$ are the predicted and GT counts for the $m^{th}$ test video, respectively.

To better evaluate the performance of different models in recognizing actions with varying periods, we expand the evaluation metrics with three variants for the OBO and MAE metrics. We split the test video set into three categories based on the average single action period length: short-, medium-, and long-period test sets. We define videos with an average single action duration of fewer than 30 frames as belonging to the short-period test set, videos with an average action duration longer than 60 frames as the long-period test set, and the remaining videos as belonging to the medium-period test set. We compute the OBO and MAE metrics on each of these sets separately.

\subsection{Implementation Details}

We implement our approach with different backbone video feature extractors respectively, including TSN~\citep{wang2016temporal}, I3D~\citep{carreira2017quo}, and ViT~\citep{dosovitskiy2021an}. For encoder-decoder Transformer, we set $\encnum=2$ and $\decnum=4$, both with 8-head attention mechanisms. The feature dimension is set to $C=512$. We set $Q=40$ queries in our method empirically. For more results with different feature channels $C$ and query numbers $Q$, please refer to the supplementary materials. The length of the video input is set to $T=512$ frames without down-sampling. We utilize the AdamW optimizer~\citep{loshchilov2017decoupled} with a learning rate of $0.002$, a batch size of $64$, and train the model for $80$ epochs. We set $\lambdahug=1.0$, $\lambdall=5.0$, $\lambdaiou=0.4$, $\lambdactrs=1.0$, confidence threshold $\alpha=0.2$. We provide further implementation details in the supplementary.

\subsection{Comparison to State-of-the-arts}
\label{sec:sota}

\begin{table*}[t]
    \centering
    \begin{minipage}[ht]{\linewidth} %
        \caption{\textbf{Generalization comparison with SOTA TRC methods on UCFRep \citep{zhang2020context} dataset.} MAE and OBO metrics for \underline{s}hort-, \underline{m}edium-, and \underline{l}ong-period actions are also reported.}
        \label{tab:result-UCFRep}
        \setlength{\tabcolsep}{4pt}
        \resizebox{\linewidth}{!}{%
        \begin{tabular}{l | c | c c | c c | c c | c c}
         \thickhline
           & Backbone & ${\rm MAE}\downarrow$ & ${\rm OBO}\uparrow$ & ${\rm MAE_{s}}\downarrow$ & ${\rm OBO_{s}}\uparrow$ & ${\rm MAE_{m}}\downarrow$ & ${\rm OBO_{m}}\uparrow$ & ${\rm MAE_{l}}\downarrow$ & ${\rm OBO_{l}}\uparrow$ \\
         \thickhline
          RepNet \citep{dwibedi2020counting} & ResNet & {0.5336} & 0.2984 & {0.6219} & {0.1739} & \underline{0.4825} & 0.3600 & {0.4996} & 0.5000 \\
          
          TransRAC \citep{hu2022transrac} & ViT & 0.6180 & {0.3143} & 0.6296 & \underline{0.1951} & 0.5842 & \underline{0.4250} & 0.6784 & 0.4118 \\
          Li \etal \citep{li2024repetitive} & 3D-ResNext & \underline{0.5227} & 0.3500 & - & - & - & - & - & - \\
          \thickhline
          \rowcolor{mygray}
          Ours & TSN & 0.6016 & 0.2959 & 0.7069 & 0.0488 & 0.5777 & \underline{0.4250} & \textbf{0.4039} & {0.5882} \\
          \rowcolor{mygray}
          Ours& I3D & \textbf{0.5194} & \underline{0.3980} & \textbf{0.4945} & \textbf{0.2195} & 0.5865 & \underline{0.4250} & \underline{0.4216} & \textbf{0.7647} \\
          \rowcolor{mygray}
          Ours & ViT & {0.5435} & \textbf{0.4184} & \underline{0.5657} & \underline{0.1951} & \textbf{0.4625} & \textbf{0.5500} & 0.6804 & \underline{0.6471} \\
          \thickhline
        \end{tabular}}
    \end{minipage}
    \vspace{-0.3cm}
\end{table*}

\begin{table*}[h]
    \centering
    \captionsetup{labelfont={color=black}}\caption{\revision{\textbf{Comparison of computational complexity and inference time across different models.} We evaluate the complexity and efficiency on varying frame lengths ($T$). OOM denotes ``Out-of-Memory'', indicating that the model ran out of memory on a 32G GPU, and therefore, no inference time could be reported.}}
    \label{tab:efficiency}
    \setlength{\tabcolsep}{16pt}
    \resizebox{0.95\linewidth}{!}{%
    \begin{tabular}{l | c | c  c c}
    \thickhline
    Model & $T$ & Params (M) & FLOPs (G) & Inference Time (s) \\
    \thickhline
    \multirow{2}{*}{RepNet \citep{dwibedi2020counting}} 
    & 64 & 43.98 & 163.44 & 3.62 \\
    & 512 & 48.94 & 6832.50 & OOM \\
    \hline
    \multirow{2}{*}{TransRAC \citep{hu2022transrac}} 
    & 64 & 42.28 & 582.02 & 2.43 \\
    & 512 & 48.22 & 8555.78 & OOM \\
    \thickhline
    \rowcolor{mygray}
    & 64 & 42.38 & 72.48 & 0.15 \\
    \rowcolor{mygray}
    \multirow{-2}{*}{Ours} & 512 & 42.38 & 574.38 & 0.78 \\
    \thickhline
    \end{tabular}}
\end{table*}

\noindent\textbf{Results on RepCountA dataset.} We compare our proposed approach to the state-of-the-art methods on the RepCountA \citep{hu2022transrac} dataset following previous work \citep{hu2022transrac, li2024repetitive} in \cref{tab:result-repcountA}. We compare with the SOTA action recognition~\citep{feichtenhofer2020x3d, liu2021tam, liu2022video}, action segmentation~\citep{huang2020improving} methods \textbf{(top block)}, and TRC~\citep{dwibedi2020counting, zhang2020context, hu2022transrac, li2024repetitive} approaches \textbf{(second block)}. We further adapt recent DETR-style action detection approaches~\citep{liu2022end, zhang2022actionformer, shi2022react} for the TRC task \textbf{(third block)}. We change their output layers accordingly and train them on RepCountA.

As shown in \cref{tab:result-repcountA}, our approach significantly outperforms the state-of-the-art methods across actions of varying lengths.
Specifically, the similarity-matrix-based methods~\citep{dwibedi2020counting, hu2022transrac,li2024repetitive} suffer from quadratic computation complexity. To manage this, they employ a sparse sampling strategy, ensuring reasonable content coverage within a limited temporal context window. However, this approach results in inferior performance for short, rapid action instances, as it tends to overlook the cycles. 
Conversely, the action detection approaches \citep{liu2022end, zhang2022actionformer, shi2022react} are primarily developed for detecting action instances specific to particular classes. To adapt them to the TRC task, we modify their output layer to incorporate class-agnostic supervision accordingly. However, relying solely on class-agnostic supervision does not support them to dynamically identify repetitive cycles based on the input videos. As a result, their performance on long, slow action instances is generally inferior compared to dedicated TRC approaches.
In contrast, our approach effectively balances the detection of actions at various speeds.

\vspace{1em}
\noindent\textbf{Results on UCFRep dataset.} We also evaluate the generalization ability of our method. Following previous work \citep{hu2022transrac,li2024repetitive}, we evaluate the model trained on the RepCountA dataset \citep{hu2022transrac} on UCFRep~\citep{zhang2020context} test set. \cref{tab:result-UCFRep} shows that our approach generally outperforms existing works, with more significant improvements observed in longer-period actions. The results demonstrate the effectiveness and generalization capability of our method.

\begin{figure*}[ht!]
    \centering
    \begin{minipage}[t]{0.97\linewidth}
        \includegraphics[width=\linewidth]{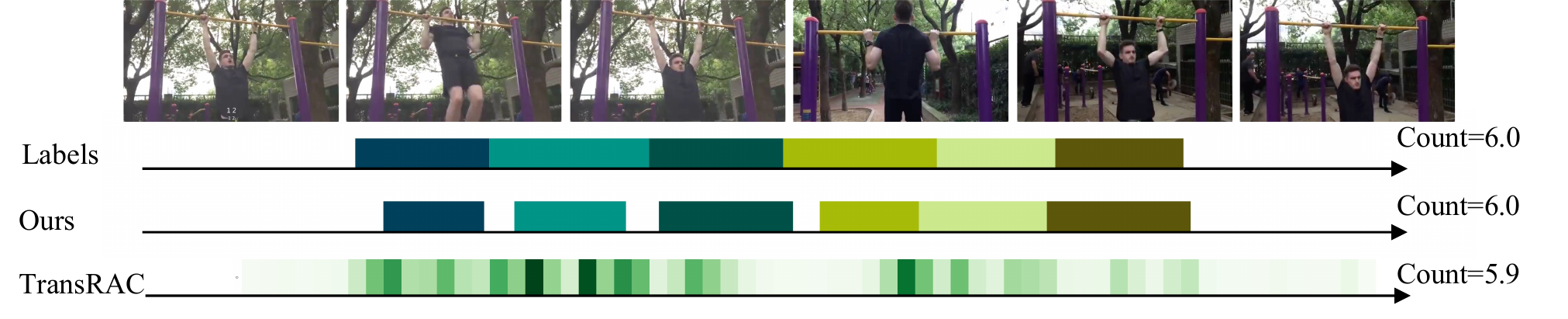}
    \end{minipage}
    \begin{minipage}[t]{0.97\linewidth}
        \includegraphics[width=1\linewidth]{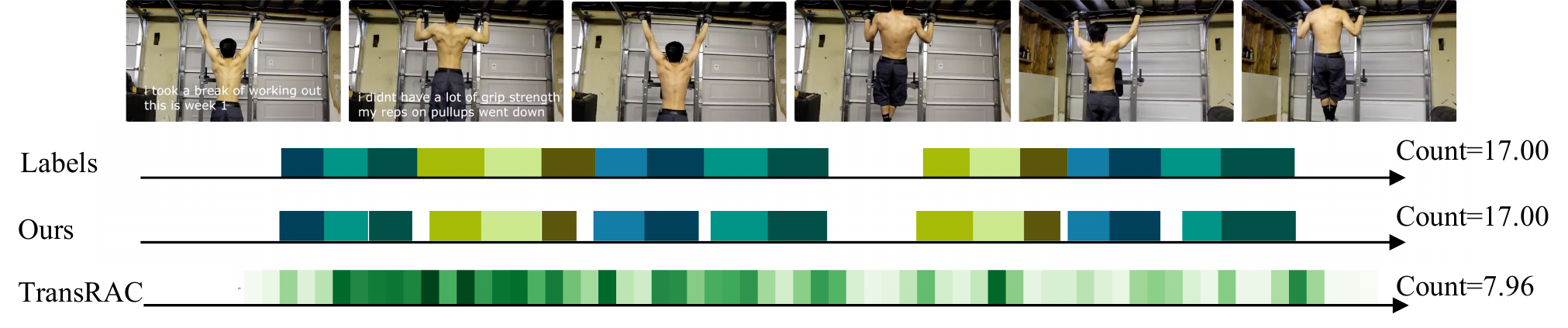} 
    \end{minipage} \\
    \caption{\textbf{Qualitative results on RepCountA dataset}. Each colored block represents a GT or predicted action instance. TransRAC represents the results by density map, and the final count value is obtained by summing the values in the density map. }
    \label{fig:result-repcountA}
\end{figure*}

\begin{figure*}[t!]
    \centering
    \begin{minipage}[t]{0.97\linewidth}
        \includegraphics[width=1\linewidth]{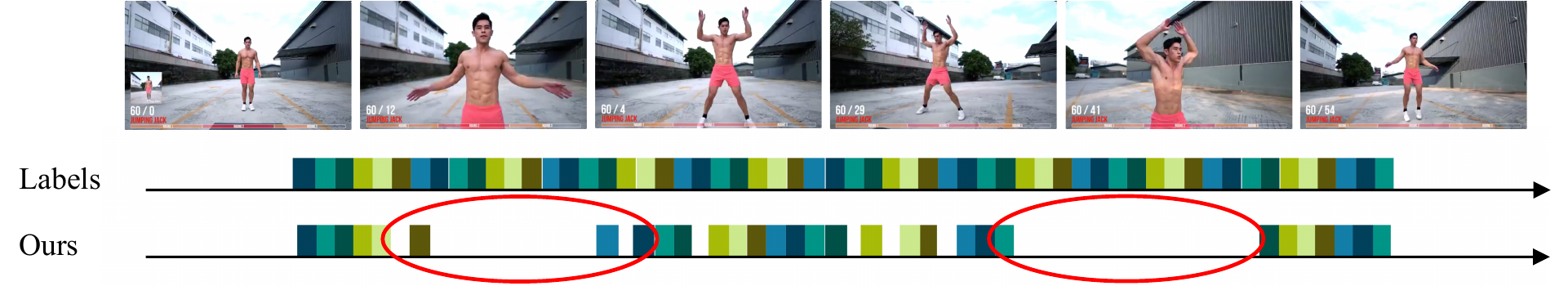} 
    \end{minipage} 
    \caption{\textbf{Visualization of failure case on RepCountA \citep{hu2022transrac} dataset.} Due to the excessive zooming in, the legs of the human body are truncated, making a large difference in the action motion feature, and resulting in several missed cycle counts. Each colored block represents a GT or predicted action instance. }
    \label{fig:failure-case}
    \vspace{-0.3cm}
\end{figure*}

\begin{figure*}[h!]
    \centering
    \begin{minipage}[t]{0.97\linewidth}
        \includegraphics[width=1\linewidth]{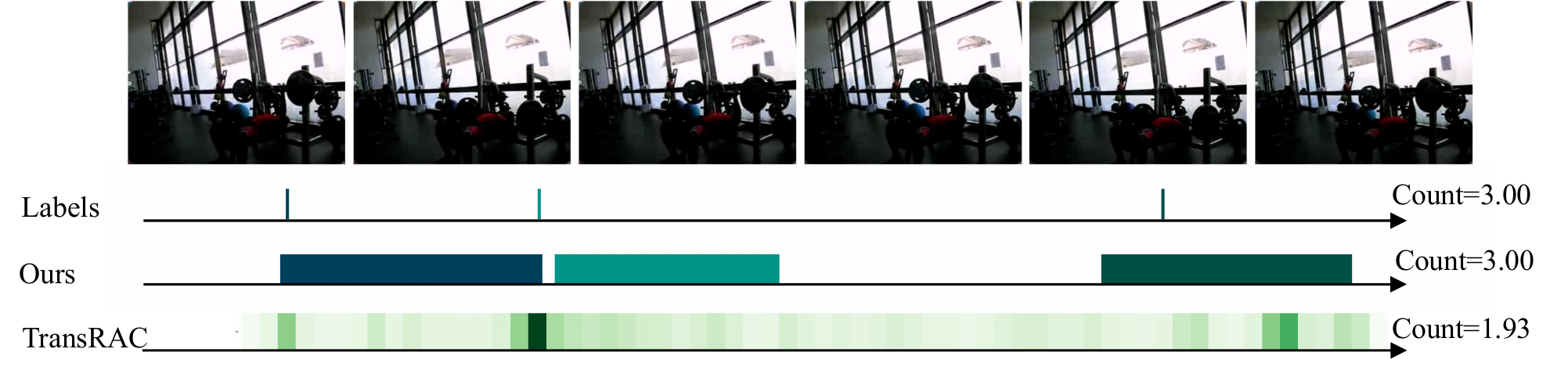} 
    \end{minipage} \\
    \begin{minipage}[t]{0.97\linewidth}
        \includegraphics[width=1\linewidth]{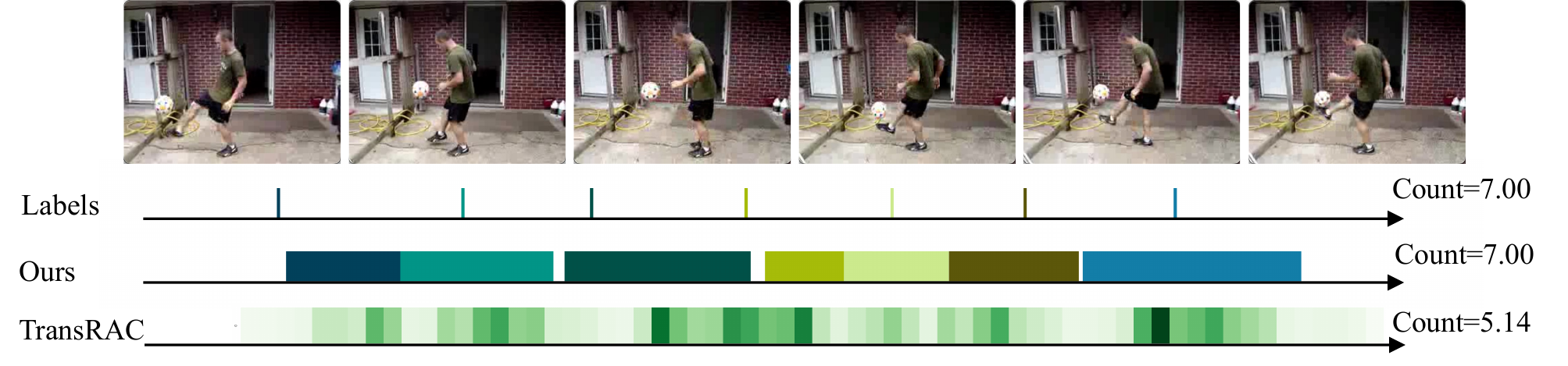}
    \end{minipage}
    \caption{\textbf{Qualitative results on UCFRep dataset}. Each colored block represents a GT or predicted action instance. TransRAC represents the results by density map, and the final count value is obtained by summing the values in the density map. The vertical lines in the labels represent the time points at which the actions begin since only the starting point annotations are provided in UCFRep \citep{zhang2020context}.
    }
    \label{fig:result-ucfrep}
    \vspace{-0.3cm}
\end{figure*}

\begin{table*}[t]
    \centering
    \begin{minipage}[t]{1\linewidth} %
        \centering
        \caption[]{\textbf{Effect of DAQ and ICL modules on RepCountA \citep{hu2022transrac} dataset}. (a) ablates DAQ strategy. (b) ablates the ICL strategy. (c) is our full model.}
        \label{tab:result-ablation}
        \vspace{-0.2cm}
        \setlength{\tabcolsep}{8pt}
        \resizebox{\linewidth}{!}{%
        \begin{tabular}{l | c c | c c | c c | c c}
         \thickhline
           & ${\rm MAE}\downarrow$ & ${\rm OBO}\uparrow$ & ${\rm MAE_{s}}\downarrow$ & ${\rm OBO_{s}}\uparrow$ & ${\rm MAE_{m}}\downarrow$ & ${\rm OBO_{m}}\uparrow$ & ${\rm MAE_{l}}\downarrow$ & ${\rm OBO_{l}}\uparrow$ \\
         \thickhline
          (a) \textit{w/o} DAQ & \underline{0.3542} & \underline{0.4172} & 0.2624 & \textbf{0.2093} & \underline{0.2515} &  \underline{0.5227} &  \underline{0.4864} & 0.4844 \\
          (b) \textit{w/o} ICL & 0.4035 & 0.4040 & \underline{0.2448} & \textbf{0.2093} & 0.3106 & 0.4545 & 0.5740 &  \underline{0.5000} \\
          \rowcolor{mygray}
          (c) Ours (full) & \textbf{0.2809} & \textbf{0.4570} & \textbf{0.2411} & \underline{0.1628} & \textbf{0.1792} & \underline{0.5455} & \textbf{0.3776} & \textbf{0.5938} \\
          \thickhline
        \end{tabular}}
    \end{minipage}
    % \vspace{0.1cm}
\end{table*}

\vspace{1em}
\noindent\textbf{Efficiency.} Additionally, we evaluate the computational complexity of our method in comparison to state-of-the-art methods. We benchmark all the methods with inference on the RepCountA dataset. \revision{As shown in \cref{tab:efficiency}, we present a detailed comparison including parameter count, FLOPs, and inference time, all measured on a single NVIDIA V100 32G GPU with a bacth size to be 1. Each method's performance across two varying frame lengths ($T=64$ and $T=512$) is evaluated on the \textbf{full model}, to provide a more comprehensive assessment.}

\revision{Notably, prior methods like RepNet \citep{dwibedi2020counting} and TransRAC \citep{hu2022transrac} were originally designed to accept video inputs primarily at $T=64$ frames. When we attempted to process $T=512$ frames with their implementations, their computational complexity increased drastically. This is because both methods rely on a temporal similarity correlation matrix, causing their complexity to grow quadratically with the input length $T$. Consequently, they encountered ``Out-of-Memory (OOM)'' errors during execution, failing to obtain their inference time.}

\revision{In contrast, our method is significantly more lightweight and achieves demonstrably faster inference times compared to prior approaches. More importantly, its computational complexity scales \textbf{linearly} with increasing input video length ($T$), which robustly ensures the algorithm's efficiency and allows it to maintain a very fast processing speed even when processing $T=512$ frames, as evidenced in \cref{tab:efficiency}.
}

\revision{Please refer to Supplementary for a decomposed model component complexity analysis.}

\subsection{Qualitative Results}

We visualize the predictions of our approach and baseline method TransRAC \citep{hu2022transrac} on the RepCountA \citep{hu2022transrac} dataset in \cref{fig:result-repcountA}. 
TransRAC \citep{hu2022transrac} represents action cycles using density maps, and the final count value is obtained by summing the values in the density map. However, this approach suffers from a lack of interpretability and finally results in miscounting. In contrast, our approach not only produces an accurate final count but also correctly localizes the action start and end positions (colored blocks) in most cases. In addition, our method exhibits robustness to changes in viewpoint, background noise, and sudden interruptions, \eg, in the second case of \cref{fig:result-repcountA}, we accurately estimate the time positions even with viewpoint changes as well as a sudden interruption in the middle of the timeline. Please refer to our supplementary video for more qualitative results. We present the video result of the first case in \cref{fig:result-repcountA}, which shows that while the subject performs pull-ups, there are moments of talking and arm waving. Our method selectively counts only the pull-up actions, as they are the primary focus and consistent with the main action instances of the video, excluding dissimilar actions like ``waving'', validating the robustness and generalization ability of our method.

In addition, our query-based representation offers excellent interpretability, making it easy to identify the issues in case of failures. For example, \cref{fig:failure-case} shows a typical failure case. Due to the excessive zooming in, the legs of the human body are truncated, making a large difference in the action motion feature, and resulting in several missed cycle counts.

We further illustrate the generalization performance of the proposed method on the unseen UCFRep \citep{zhang2020context} test set in \cref{fig:result-ucfrep}. We directly apply our trained model and do not use UCFRep training data. Our model still accurately recognizes the action instances and gets the correct count in the challenging cases, indicating robust generalization ability. Specifically, the top case exhibits extreme viewpoint and lighting conditions, while the bottom case contains the action of soccer juggling which is not seen in the training set.  We attribute the performance advantage to the proposed DAQ and ICL designs, which empower the model to adaptively adjust the action queries based on the input video features and effectively localize similar (repetitive) action instances, distinguishing them from the background noise actions.

\subsection{Ablation Study}
\label{sec:ablation}
\noindent\textbf{Effect of DAQ and ICL.} 
We implement two ablated models to study the efficacy of the proposed DAQ and ICL designs. \cref{tab:result-ablation} presents the results on RepCountA \citep{hu2022transrac} using the TSN backbone. In ablation (a), we substitute the proposed DAQ module with a static action query, where the action queries $\queryact$ are also learnable variables, instead of using the action queries filtered out by the encoder and query selection modules. In ablation (b), we eliminate the ICL design among the action queries. The results demonstrate that both DAQ and ICL contribute to performance improvement, particularly in terms of counting medium and long actions.

\begin{figure}[t]
    \centering
    \begin{minipage}[t]{\linewidth}
        \includegraphics[width=1\linewidth]{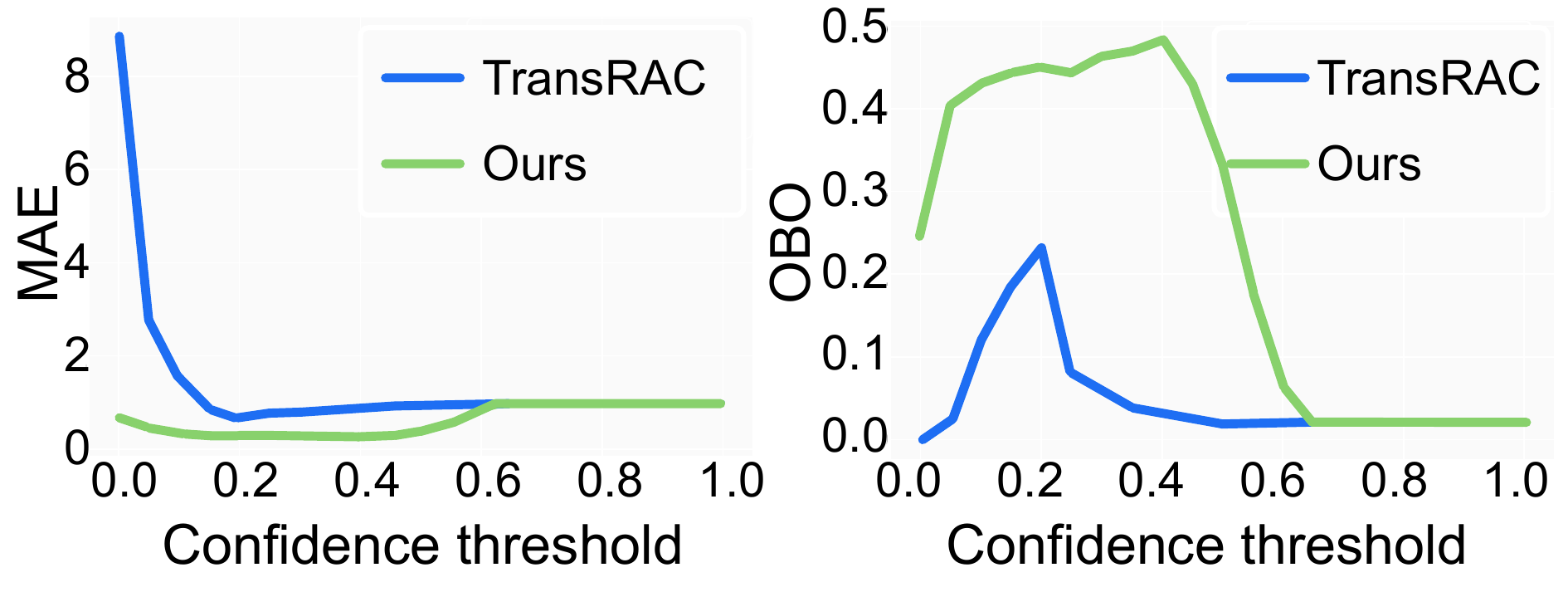} 
    \end{minipage} 
    \vspace{0.05cm}
    \caption{\textbf{Results of different confidence thresholds of our method and TransRAC \citep{hu2022transrac}.} We depict the MAE (left) and OBO (right) curves of our method (green curve) and the TransRAC (blue curve) approach concerning different confidence thresholds. The metrics of TransRAC are obtained by binarizing the density map of TransRAC output and then summing to obtain the final count.}
    \label{fig:supp-conf-thres}
\end{figure}

\vspace{1em}
\noindent\textbf{Confidence threshold $\alpha$.}
We report the impact of different confidence thresholds\revision{, which range from 0 to 1} for classifying ``repetitive actions'' on the final counts in \cref{fig:supp-conf-thres}. \revision{During inference, queries with prediction confidence scores below the threshold are classified as ``others'', while the rest are considered ``repetitive actions''.} We illustrate the MAE (left) and OBO (right) curves for our method (green curve) and the TransRAC \citep{hu2022transrac} approach (blue curve) across various confidence thresholds $\alpha$. The metrics for TransRAC are derived by binarizing its output density map and summing the results to obtain the final count. 

\revision{As the figure illustrates, our model performance drops significantly when the threshold exceeds 0.5, \eg OBO becomes nearly zero, since most queries are mistakenly filtered out. This aligns with our expectations. Conversely, when the threshold is set between 0.3 and 0.5, the model yields consistently strong performance. However, setting the threshold too low also leads to performance degradation, as nearly all queries are then classified as ``repetitive actions'', including irrelevant or noisy ones. This highlights the importance of selecting an appropriate threshold range to achieve optimal performance, as the results indicate that setting the threshold within the range of $0.2$ to $0.4$ yields consistently strong performance for our method. Besides, our method consistently outperforms TransRAC \citep{hu2022transrac} (blue curve) by a considerable margin.}

\section{Conclusion}
\label{sec:conclusion}

In conclusion, we provide an innovative perspective for tackling TRC tasks, which reduces computational complexity and maintains robustness across varying action lengths. To address open-set action categories, we propose DAQ for improved generalization, and ICL for recognizing repetitive actions and distinguishing them from distractions. Integrating DAQ and ICL, our method adaptively identifies contextually similar action instances. Experimental results on challenging benchmarks demonstrate our approach's superiority over SOTAs in both accuracy and efficiency, while adeptly balancing diverse action speeds and video durations, establishing a solid foundation for practical implementations in real-world scenarios.

\backmatter

\bmhead{Acknowledgements}
This work was supported by National Science and Technology Major Project (2022ZD0114904).

\bmhead{Data Availability Statements.}
The RepCountA \citep{hu2022transrac} and UCFRep \citep{zhang2020context} datasets that support the findings of this study are publicly available on GitHub at \url{https://github.com/SvipRepetitionCounting/TransRAC} and \url{https://github.com/Xiaodomgdomg/Deep-Temporal-Repetition-Counting}, respectively.

\bibliography{sn-bibliography}% common bib file

@string {CVPR = "Conference on Computer Vision and Pattern Recognition (CVPR)"}

@string {ICCV = "International Conference on Computer Vision (ICCV)"}

@string {BMVC = "British Machine Vision Conference (BMVC)"}

@inproceedings{li2024repetitive,
  title={Repetitive Action Counting With Motion Feature Learning},
  author={Li, Xinjie and Xu, Huijuan},
  booktitle={Proceedings of the IEEE/CVF Winter Conference on Applications of Computer Vision},
  pages={6499--6508},
  year={2024}
}

@article{kuhn1955hungarian,
  title={The Hungarian method for the assignment problem},
  author={Kuhn, Harold W},
  journal={Naval research logistics quarterly},
  volume={2},
  number={1-2},
  pages={83--97},
  year={1955},
  publisher={Wiley Online Library}
}

@inproceedings{zhang2022actionformer,
  title={ActionFormer: Localizing Moments of Actions with Transformers},
  author={Zhang, Chen-Lin and Wu, Jianxin and Li, Yin},
  booktitle={European Conference on Computer Vision},
  series={LNCS},
  volume={13664},
  pages={492-510},
  year={2022}
}

@inproceedings{liu2021tam,
  title={Tam: Temporal adaptive module for video recognition},
  author={Liu, Zhaoyang and Wang, Limin and Wu, Wayne and Qian, Chen and Lu, Tong},
  booktitle=ICCV,
  pages={13708--13718},
  year={2021}
}

@inproceedings{feichtenhofer2020x3d,
  title={X3d: Expanding architectures for efficient video recognition},
  author={Feichtenhofer, Christoph},
  booktitle={Proceedings of the IEEE/CVF conference on computer vision and pattern recognition},
  pages={203--213},
  year={2020}
}

@inproceedings{
dosovitskiy2021an,
title={An Image is Worth 16x16 Words: Transformers for Image Recognition at Scale},
author={Alexey Dosovitskiy and Lucas Beyer and Alexander Kolesnikov and Dirk Weissenborn and Xiaohua Zhai and Thomas Unterthiner and Mostafa Dehghani and Matthias Minderer and Georg Heigold and Sylvain Gelly and Jakob Uszkoreit and Neil Houlsby},
booktitle={International Conference on Learning Representations},
year={2021},
}

@inproceedings{laptev2005periodic,
  title={Periodic motion detection and segmentation via approximate sequence alignment},
  author={Laptev, Ivan and Belongie, Serge J and P{\'e}rez, Patrick and Wills, Josh},
  booktitle={Tenth IEEE International Conference on Computer Vision (ICCV'05) Volume 1},
  volume={1},
  pages={816--823},
  year={2005},
  organization={IEEE}
}

@inproceedings{huang2020improving,
  title={Improving action segmentation via graph-based temporal reasoning},
  author={Huang, Yifei and Sugano, Yusuke and Sato, Yoichi},
  booktitle={Proceedings of the IEEE/CVF conference on computer vision and pattern recognition},
  pages={14024--14034},
  year={2020}
}

@inproceedings{carreira2017quo,
  title={Quo vadis, action recognition? a new model and the kinetics dataset},
  author={Carreira, Joao and Zisserman, Andrew},
  booktitle={proceedings of the IEEE Conference on Computer Vision and Pattern Recognition},
  pages={6299--6308},
  year={2017}
}

@inproceedings{liu2022video,
  title={Video swin transformer},
  author={Liu, Ze and Ning, Jia and Cao, Yue and Wei, Yixuan and Zhang, Zheng and Lin, Stephen and Hu, Han},
  booktitle={Proceedings of the IEEE/CVF conference on computer vision and pattern recognition},
  pages={3202--3211},
  year={2022}
}

@article{cutler2000robust,
  title={Robust real-time periodic motion detection, analysis, and applications},
  author={Cutler, Ross and Davis, Larry S.},
  journal={IEEE Transactions on Pattern Analysis and Machine Intelligence},
  volume={22},
  number={8},
  pages={781--796},
  year={2000},
  publisher={IEEE}
}

@inproceedings{azy2008segmentation,
  title={Segmentation of periodically moving objects},
  author={Azy, Ousman and Ahuja, Narendra},
  booktitle={2008 19th International Conference on Pattern Recognition},
  pages={1--4},
  year={2008},
  organization={IEEE}
}

@inproceedings{pogalin2008visual,
  title={Visual quasi-periodicity},
  author={Pogalin, Erik and Smeulders, Arnold WM and Thean, Andrew HC},
  booktitle={2008 IEEE Conference on Computer Vision and Pattern Recognition},
  pages={1--8},
  year={2008},
  organization={IEEE}
}

@inproceedings{runia2018real,
  title={Real-world repetition estimation by div, grad and curl},
  author={Runia, Tom FH and Snoek, Cees GM and Smeulders, Arnold WM},
  booktitle={Proceedings of the IEEE conference on computer vision and pattern recognition},
  pages={9009--9017},
  year={2018}
}

@inproceedings{dwibedi2020counting,
  title={Counting out time: Class agnostic video repetition counting in the wild},
  author={Dwibedi, Debidatta and Aytar, Yusuf and Tompson, Jonathan and Sermanet, Pierre and Zisserman, Andrew},
  booktitle={Proceedings of the IEEE/CVF conference on computer vision and pattern recognition},
  pages={10387--10396},
  year={2020}
}

@inproceedings{hu2022transrac,
  title={TransRAC: Encoding Multi-scale Temporal Correlation with Transformers for Repetitive Action Counting},
  author={Hu, Huazhang and Dong, Sixun and Zhao, Yiqun and Lian, Dongze and Li, Zhengxin and Gao, Shenghua},
  booktitle={Proceedings of the IEEE/CVF Conference on Computer Vision and Pattern Recognition},
  pages={19013--19022},
  year={2022}
}

@inproceedings{levy2015live,
  title={Live repetition counting},
  author={Levy, Ofir and Wolf, Lior},
  booktitle={Proceedings of the IEEE international conference on computer vision},
  pages={3020--3028},
  year={2015}
}

@inproceedings{zhang2020context,
  title={Context-aware and scale-insensitive temporal repetition counting},
  author={Zhang, Huaidong and Xu, Xuemiao and Han, Guoqiang and He, Shengfeng},
  booktitle={Proceedings of the IEEE/CVF Conference on Computer Vision and Pattern Recognition},
  pages={670--678},
  year={2020}
}

@article{zhu2023human,
  title={Human motion generation: A survey},
  author={Zhu, Wentao and Ma, Xiaoxuan and Ro, Dongwoo and Ci, Hai and Zhang, Jinlu and Shi, Jiaxin and Gao, Feng and Tian, Qi and Wang, Yizhou},
  journal={IEEE Transactions on Pattern Analysis and Machine Intelligence},
  year={2023},
  publisher={IEEE}
}

@inproceedings{carion2020end,
  title={End-to-end object detection with transformers},
  author={Carion, Nicolas and Massa, Francisco and Synnaeve, Gabriel and Usunier, Nicolas and Kirillov, Alexander and Zagoruyko, Sergey},
  booktitle={Computer Vision--ECCV 2020: 16th European Conference, Glasgow, UK, August 23--28, 2020, Proceedings, Part I 16},
  pages={213--229},
  year={2020},
  organization={Springer}
}

@inproceedings{shi2022react,
  title={React: Temporal action detection with relational queries},
  author={Shi, Dingfeng and Zhong, Yujie and Cao, Qiong and Zhang, Jing and Ma, Lin and Li, Jia and Tao, Dacheng},
  booktitle={Computer Vision--ECCV 2022: 17th European Conference, Tel Aviv, Israel, October 23--27, 2022, Proceedings, Part X},
  pages={105--121},
  year={2022},
  organization={Springer}
}

@inproceedings{chao2018rethinking,
  title={Rethinking the faster r-cnn architecture for temporal action localization},
  author={Chao, Yu-Wei and Vijayanarasimhan, Sudheendra and Seybold, Bryan and Ross, David A and Deng, Jia and Sukthankar, Rahul},
  booktitle={Proceedings of the IEEE conference on computer vision and pattern recognition},
  pages={1130--1139},
  year={2018}
}

@inproceedings{li2021three,
  title={Three birds with one stone: Multi-task temporal action detection via recycling temporal annotations},
  author={Li, Zhihui and Yao, Lina},
  booktitle={Proceedings of the IEEE/CVF Conference on Computer Vision and Pattern Recognition},
  pages={4751--4760},
  year={2021}
}

@inproceedings{lin2021learning,
  title={Learning salient boundary feature for anchor-free temporal action localization},
  author={Lin, Chuming and Xu, Chengming and Luo, Donghao and Wang, Yabiao and Tai, Ying and Wang, Chengjie and Li, Jilin and Huang, Feiyue and Fu, Yanwei},
  booktitle={Proceedings of the IEEE/CVF Conference on Computer Vision and Pattern Recognition},
  pages={3320--3329},
  year={2021}
}

@inproceedings{qing2021temporal,
  title={Temporal context aggregation network for temporal action proposal refinement},
  author={Qing, Zhiwu and Su, Haisheng and Gan, Weihao and Wang, Dongliang and Wu, Wei and Wang, Xiang and Qiao, Yu and Yan, Junjie and Gao, Changxin and Sang, Nong},
  booktitle={Proceedings of the IEEE/CVF conference on computer vision and pattern recognition},
  pages={485--494},
  year={2021}
}

@inproceedings{zeng2019graph,
  title={Graph convolutional networks for temporal action localization},
  author={Zeng, Runhao and Huang, Wenbing and Tan, Mingkui and Rong, Yu and Zhao, Peilin and Huang, Junzhou and Gan, Chuang},
  booktitle=ICCV,
  pages={7094--7103},
  year={2019}
}

@inproceedings{zhao2017temporal,
  title={Temporal action detection with structured segment networks},
  author={Zhao, Yue and Xiong, Yuanjun and Wang, Limin and Wu, Zhirong and Tang, Xiaoou and Lin, Dahua},
  booktitle={Proceedings of the IEEE international conference on computer vision},
  pages={2914--2923},
  year={2017}
}

@inproceedings{redmon2016you,
  title={You only look once: Unified, real-time object detection},
  author={Redmon, Joseph and Divvala, Santosh and Girshick, Ross and Farhadi, Ali},
  booktitle={Proceedings of the IEEE conference on computer vision and pattern recognition},
  pages={779--788},
  year={2016}
}

@inproceedings{buch2019end,
  title={End-to-end, single-stream temporal action detection in untrimmed videos},
  author={Buch, Shyamal and Escorcia, Victor and Ghanem, Bernard and Fei-Fei, Li and Niebles, Juan Carlos},
  booktitle={Procedings of the British Machine Vision Conference 2017},
  year={2019},
  organization={British Machine Vision Association}
}

@inproceedings{shou2017cdc,
  title={Cdc: Convolutional-de-convolutional networks for precise temporal action localization in untrimmed videos},
  author={Shou, Zheng and Chan, Jonathan and Zareian, Alireza and Miyazawa, Kazuyuki and Chang, Shih-Fu},
  booktitle={Proceedings of the IEEE conference on computer vision and pattern recognition},
  pages={5734--5743},
  year={2017}
}

@inproceedings{yuan2017temporal,
  title={Temporal action localization by structured maximal sums},
  author={Yuan, Zehuan and Stroud, Jonathan C and Lu, Tong and Deng, Jia},
  booktitle={Proceedings of the IEEE Conference on Computer Vision and Pattern Recognition},
  pages={3684--3692},
  year={2017}
}

@article{liu2022end,
  title={End-to-end temporal action detection with transformer},
  author={Liu, Xiaolong and Wang, Qimeng and Hu, Yao and Tang, Xu and Zhang, Shiwei and Bai, Song and Bai, Xiang},
  journal={IEEE Transactions on Image Processing},
  volume={31},
  pages={5427--5441},
  year={2022},
  publisher={IEEE}
}

@inproceedings{tan2021relaxed,
  title={Relaxed transformer decoders for direct action proposal generation},
  author={Tan, Jing and Tang, Jiaqi and Wang, Limin and Wu, Gangshan},
  booktitle=ICCV,
  pages={13526--13535},
  year={2021}
}

@article{vaswani2017attention,
  title={Attention is all you need},
  author={Vaswani, Ashish and Shazeer, Noam and Parmar, Niki and Uszkoreit, Jakob and Jones, Llion and Gomez, Aidan N and Kaiser, {\L}ukasz and Polosukhin, Illia},
  journal={Advances in neural information processing systems},
  volume={30},
  year={2017}
}

@inproceedings{wang2021oadtr,
  title={Oadtr: Online action detection with transformers},
  author={Wang, Xiang and Zhang, Shiwei and Qing, Zhiwu and Shao, Yuanjie and Zuo, Zhengrong and Gao, Changxin and Sang, Nong},
  booktitle=ICCV,
  pages={7565--7575},
  year={2021}
}

@inproceedings{zhang2022dino,
title={{DINO}: {DETR} with Improved DeNoising Anchor Boxes for End-to-End Object Detection},
author={Hao Zhang and Feng Li and Shilong Liu and Lei Zhang and Hang Su and Jun Zhu and Lionel Ni and Heung-Yeung Shum},
booktitle={The Eleventh International Conference on Learning Representations },
year={2023},
}

@inproceedings{liu2022dab,
title={{DAB}-{DETR}: Dynamic Anchor Boxes are Better Queries for {DETR}},
author={Shilong Liu and Feng Li and Hao Zhang and Xiao Yang and Xianbiao Qi and Hang Su and Jun Zhu and Lei Zhang},
booktitle={International Conference on Learning Representations},
year={2022},
}

@inproceedings{zhu2020deformable,
title={Deformable DETR: Deformable Transformers for End-to-End Object Detection},
author={Xizhou Zhu and Weijie Su and Lewei Lu and Bin Li and Xiaogang Wang and Jifeng Dai},
booktitle={International Conference on Learning Representations},
year={2021},
}

@inproceedings{fieraru2021aifit,
  title={Aifit: Automatic 3d human-interpretable feedback models for fitness training},
  author={Fieraru, Mihai and Zanfir, Mihai and Pirlea, Silviu Cristian and Olaru, Vlad and Sminchisescu, Cristian},
  booktitle={Proceedings of the IEEE/CVF Conference on Computer Vision and Pattern Recognition},
  pages={9919--9928},
  year={2021}
}

@inproceedings{he2020momentum,
  title={Momentum contrast for unsupervised visual representation learning},
  author={He, Kaiming and Fan, Haoqi and Wu, Yuxin and Xie, Saining and Girshick, Ross},
  booktitle={Proceedings of the IEEE/CVF conference on computer vision and pattern recognition},
  pages={9729--9738},
  year={2020}
}

@inproceedings{wang2016temporal,
  title={Temporal segment networks: Towards good practices for deep action recognition},
  author={Wang, Limin and Xiong, Yuanjun and Wang, Zhe and Qiao, Yu and Lin, Dahua and Tang, Xiaoou and Van Gool, Luc},
  booktitle={European conference on computer vision},
  pages={20--36},
  year={2016},
  organization={Springer}
}

@inproceedings{rezatofighi2019generalized,
  title={Generalized intersection over union: A metric and a loss for bounding box regression},
  author={Rezatofighi, Hamid and Tsoi, Nathan and Gwak, JunYoung and Sadeghian, Amir and Reid, Ian and Savarese, Silvio},
  booktitle={Proceedings of the IEEE/CVF conference on computer vision and pattern recognition},
  pages={658--666},
  year={2019}
}

@article{tsai1994cyclic,
  title={Cyclic motion detection for motion based recognition},
  author={Tsai, Ping-Sing and Shah, Mubarak and Keiter, Katharine and Kasparis, Takis},
  journal={Pattern recognition},
  volume={27},
  number={12},
  pages={1591--1603},
  year={1994},
  publisher={Elsevier}
}

@inproceedings{thangali2005periodic,
  title={Periodic motion detection and estimation via space-time sampling},
  author={Thangali, Ashwin and Sclaroff, Stan},
  booktitle={2005 Seventh IEEE Workshops on Applications of Computer Vision (WACV/MOTION'05)-Volume 1},
  volume={2},
  pages={176--182},
  year={2005},
  organization={IEEE}
}

@inproceedings{chetverikov2006motion,
  title={On motion periodicity of dynamic textures.},
  author={Chetverikov, Dmitry and Fazekas, S{\'a}ndor},
  booktitle={BMVC},
  volume={1},
  pages={167--176},
  year={2006},
  organization={Citeseer}
}

@inproceedings{meng2021conditional,
  title={Conditional detr for fast training convergence},
  author={Meng, Depu and Chen, Xiaokang and Fan, Zejia and Zeng, Gang and Li, Houqiang and Yuan, Yuhui and Sun, Lei and Wang, Jingdong},
  booktitle=ICCV,
  pages={3651--3660},
  year={2021}
}

@article{loshchilov2017decoupled,
  title={Decoupled weight decay regularization},
  author={Loshchilov, Ilya and Hutter, Frank},
  journal={arXiv preprint arXiv:1711.05101},
  year={2017}
}

@article{soomro2012ucf101,
  title={UCF101: A dataset of 101 human actions classes from videos in the wild},
  author={Soomro, Khurram and Zamir, Amir Roshan and Shah, Mubarak},
  journal={arXiv preprint arXiv:1212.0402},
  year={2012}
}

@inproceedings{lin2019bmn,
  title={Bmn: Boundary-matching network for temporal action proposal generation},
  author={Lin, Tianwei and Liu, Xiao and Li, Xin and Ding, Errui and Wen, Shilei},
  booktitle=ICCV,
  pages={3889--3898},
  year={2019}
}

@inproceedings{he2016deep,
  title={Deep residual learning for image recognition},
  author={He, Kaiming and Zhang, Xiangyu and Ren, Shaoqing and Sun, Jian},
  booktitle=CVPR,
  year={2016}
}

\end{document}